\documentclass[conference]{IEEEtran}
\IEEEoverridecommandlockouts
\usepackage{subcaption}
\usepackage{cite}
\usepackage{amsmath,amssymb,amsfonts}
\usepackage{algorithmic}
\usepackage{graphicx}
\usepackage{textcomp}
\usepackage{xcolor}
\usepackage{multirow}
\usepackage{multicol}
\usepackage[T1]{fontenc}
\usepackage{aecompl}
\def\BibTeX{{\rm B\kern-.05em{\sc i\kern-.025em b}\kern-.08em
    T\kern-.1667em\lower.7ex\hbox{E}\kern-.125emX}

}
	
\PassOptionsToPackage{bookmarks=false}{hyperref}

\begin{document}

\title{Robust and Efficient Memory Network for\\ Video Object Segmentation\\
\thanks{This work was partially supported by the National Natural Science Foundation of China (Grant Nos. 61802197, 62072449, and 61632003), the Science and Technology Development Fund, Macau SAR (Grant Nos. 0018/2019/AKP
	and SKL-IOTSC(UM)-2021-2023), the Guangdong Science and Technology Department (Grant No. 2020B1515130001), and University of Macau (Grant No. MYRG2020-00253-FST, MYRG2022-00059-FST).}
}


\author{\IEEEauthorblockN{1\textsuperscript{st} Yadang Chen}
\IEEEauthorblockA{\textit{School of Computer Science} \\
\textit{Nanjing University of Information Science and
	Technology}\\
Nanjing, China \\
}
\\
\IEEEauthorblockN{3\textsuperscript{rd} Zhi-Xin Yang}
\IEEEauthorblockA{\textit{State Key Laboratory of Internet of Things for Smart City} \\
\textit{ University of Macau}\\
Macau, China\\
}
\and
\IEEEauthorblockN{2\textsuperscript{nd} Dingwei Zhang}
\IEEEauthorblockA{\textit{School of Software} \\
\textit{Nanjing University of Information Science and
	Technology}\\
Nanjing, China \\
}
\\
\IEEEauthorblockN{4\textsuperscript{th} Enhua Wu}
\IEEEauthorblockA{\textit{State Key Laboratory of Computer Science} \\
\textit{University of Chinese Academy of Sciences}\\
Beijing, China \\
}
}

\maketitle

\begin{abstract}
This paper proposes a Robust and Efficient Memory Network, referred to as REMN, for studying semi-supervised video object segmentation (VOS). Memory-based methods have recently achieved outstanding VOS performance by performing non-local pixel-wise matching between the query and memory. However, these methods have two limitations. 1) Non-local matching could cause distractor objects in the background to be incorrectly segmented. 2) Memory features with high temporal redundancy consume significant computing resources. For limitation 1, we introduce a local attention mechanism that tackles the background distraction by enhancing the features of foreground objects with the previous mask. For limitation 2, we first adaptively decide whether to update the memory features depending on the variation of foreground objects to reduce temporal redundancy. Second, we employ a dynamic memory bank, which uses a lightweight and differentiable soft modulation gate to decide how many memory features need to be removed in the temporal dimension. Experiments demonstrate that our REMN achieves state-of-the-art results on DAVIS 2017, with a $\mathcal{J\&F}$ score of 86.3\% and on YouTube-VOS 2018, with a $\mathcal{G}$ over mean of 85.5\%. Furthermore, our network shows a high inference speed of 25+ FPS and uses relatively few computing resources.
\end{abstract}


\section{Introduction}
\label{sec:intro}

Video object segmentation (VOS) is a challenging task in computer vision and has attracted much attention in autonomous driving and video editing. The critical issue in VOS, as opposed to image segmentation, is to make full use of the spatio-temporal cues between frames. In this paper, we focus on semi-supervised VOS, which is to segment the foreground object in the remaining video sequence, given the annotation of the foreground object in the first frame.

Recently, a memory-based method proposed by the Space-Time Memory network (STM)\cite{oh2019video} has been widely used in semi-VOS. STM and its variants\cite{liang2020video,seong2020kernelized,chen2019appearance,xie2021efficient,yang2020collaborative,li2020fast,cheng2021modular,chen2022fast,chen2019multilevel,hu2021learning,seong2021hierarchical,Zhou2019,liang2021video,park2022per,wang2021swiftnet,lin2022swem} perform dot product operation to apply spatio-temporal matching between the query frame (i.e., current frame) and templates in the memory bank, which are features of the memory frames (i.e., first and the processed frames). However, these methods employ dot product operations with low memory coverage and need to construct a memory bank for each object to store the features, which results in inefficient segmentation. To this end, GC\cite{li2020fast} develops a compact global representation to reduce computational complexities. Also, Space-Time Correspondence Network (STCN)\cite{cheng2021rethinking} uses more efficient  L2 similarity in place of the dot product operation to compute target-agnostic frame-frame correspondences that share a single memory bank for all objects.
\begin{figure}[!t]
	\centering
	\subfloat[]{\includegraphics[width=0.245\textwidth, height=1.15in]{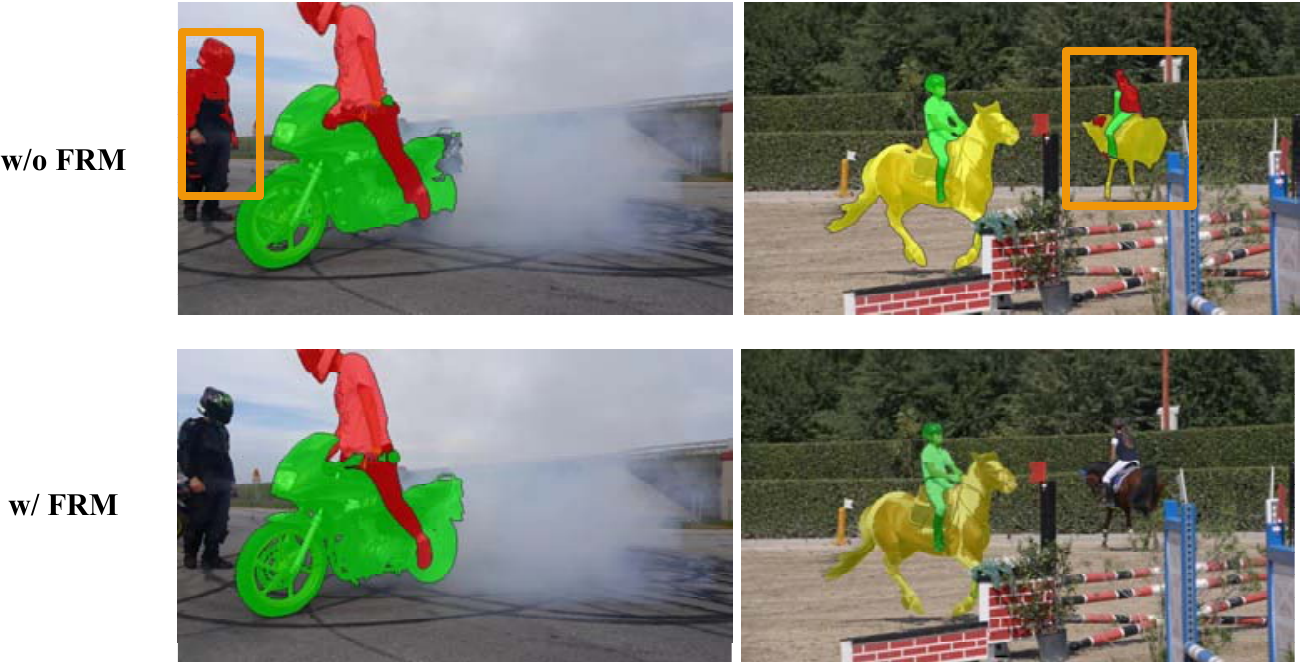}%
		\label{issues-1}}
	\hfill
	\subfloat[]{\includegraphics[width=0.24\textwidth,height=1.15in]{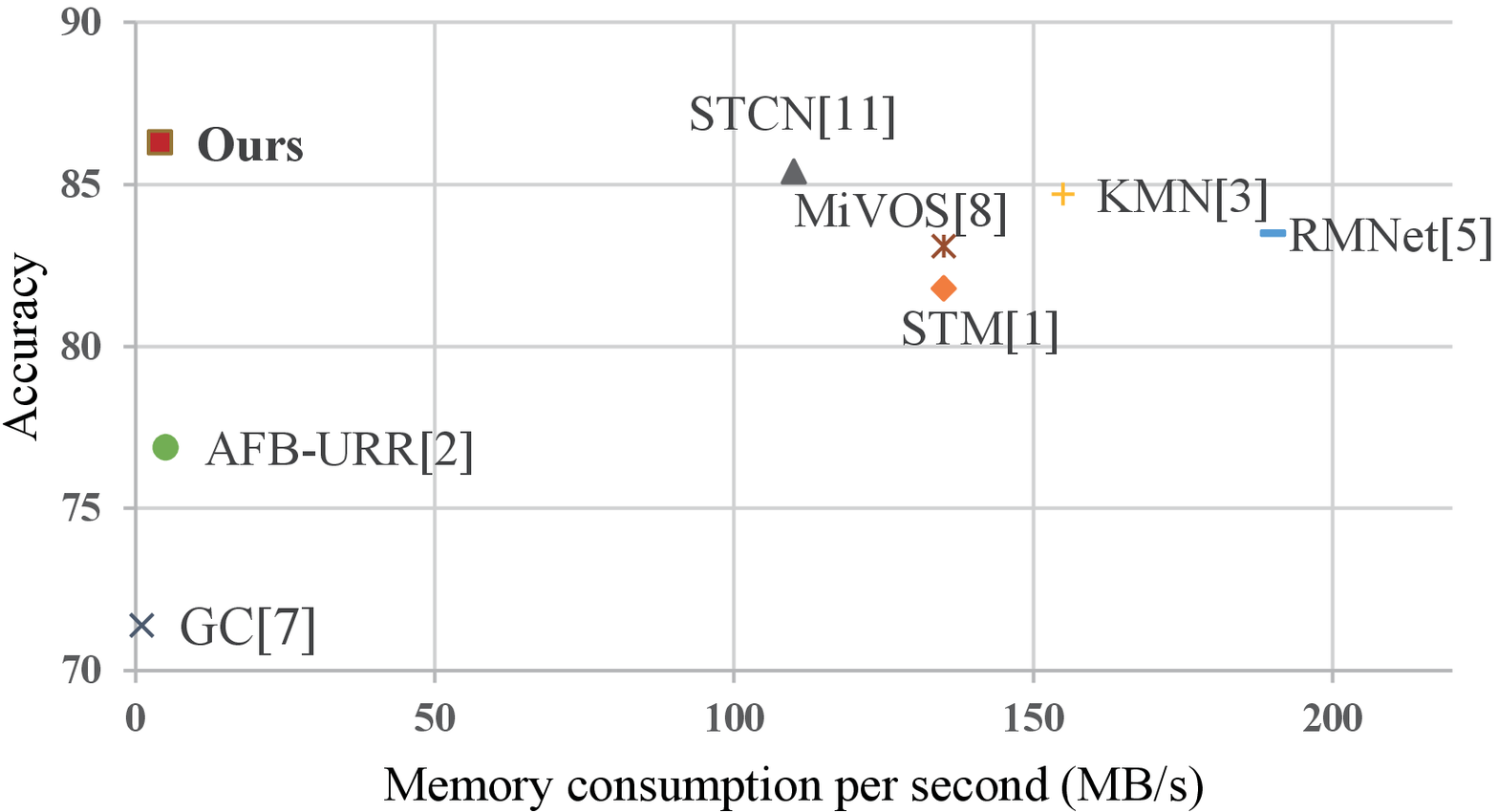}%
		\label{issues-2}}
	\caption{Two limitations in memory-based methods. (a) Background objects similar to the foreground are incorrectly segmented, as shown in the first row. The results in the second row show that our proposed Foreground Reinforcing Module (FRM) can effectively suppress background distraction. (b) Accuracy and  memory consumption of memory-based methods on DAVIS 2017 validation set. Most methods compromise computing resources for better performance.} 
	\label{issues}
\end{figure}

Although the above methods have achieved state-of-the-art performance in the VOS benchmark\cite{pont20172017,xu2018youtube}, two limitations remain unresolved. First, as shown in Fig. \ref{issues}(a), distractor objects in the background are incorrectly segmented as foreground. The reason is that both the dot product used by STM and the L2 similarity proposed by STCN perform non-local pixel-wise matching without considering temporal consistency. In general, the foreground and surrounding pixels are more valuable in the matching, while other pixels have less contribution. Although some methods\cite{seong2020kernelized,xie2021efficient,yang2020collaborative,cho2022tackling,zhang2020spatial} constrain the matching space to mitigate background distraction by bi-directional matching (e.g., KMN\cite{seong2020kernelized}, CFBI\cite{yang2020collaborative}) or optical flow models (e.g., RMNet\cite{xie2021efficient}), they extremely introduce many parameters that increase the computational burden. Second, many features with high temporal redundancy are stored in the memory bank, resulting in extensive usage of computing resources. As illustrated in Fig. \ref{issues}(b), most methods\cite{seong2020kernelized,oh2019video,xie2021efficient,cheng2021modular,cheng2021rethinking} compromise computing resources for better performance. Even though they employ a fixed sampling interval, there is still a high temporal redundancy between memory features. While AFB-URR\cite{liang2020video} discards old features according to the least frequently used index and GC\cite{li2020fast} keeps a fixed-sized set of features by global context, they will lose the critical features when the foreground objects are occluded or deformed.

We address the above limitations by introducing the following innovations in our Robust and Efficient Memory Network (REMN). For the first limitation, we propose a Foreground Reinforcing Module (FRM) to utilize the spatial location prior provided by the previous frame's mask. Unlike previous methods\cite{seong2020kernelized,xie2021efficient,yang2020collaborative,zhang2020spatial}, FRM performs a more lightweight local attention operation to enhance foreground features in the query frame, rather than introducing an additional model to constrain the matching space. The attention weights in the map indicate the spatial correlation between each object pixel in the previous frame and pixels within the local region at the corresponding position in the query frame. Fig. \ref{issues}(a) demonstrates the effectiveness of our FRM. For the second limitation, we first introduce an Adaptive Sampling Module (ASM), which decides the sampling interval by computing the difference in shape and position of the object mask between the query frame and the latest memory frame. Although ASM can reduce temporal redundancy somewhat by controlling the sample interval, temporal redundancy still exists as the video duration increases. Therefore, we also employ a dynamic memory bank consisting of a Redundancy Reduction Module (RRM). Instead of naively discarding obsolete features, RRM adaptively reduces feature redundancy based on the temporal policy of soft modulation gate output. The feature distribution of RRM is similar to that of the original to avoid losing critical features. As a result, even with very long videos, our network also keeps a fixed-size memory bank for inference at a fast speed.

Our contributions can be summarized as follows:
\begin{itemize}
	\item {We introduce a lightweight Foreground Reinforcing Module to suppress background distraction caused by non-local pixel-wise matching.}
	\item {We present an efficient Adaptive Sampling Module and Redundancy Reduction Module to reduce the temporal redundancy of memory features.}
	\item {Our REMN achieves state-of-the-art performance on DAVIS 2017 and YouTube-VOS 2018 using relatively few computing resources and maintains a high inference speed of 25+ FPS.}
\end{itemize}

\section{Proposed Method}
\begin{figure*}[htbp]
	\centering
	\includegraphics[width=0.98\textwidth]{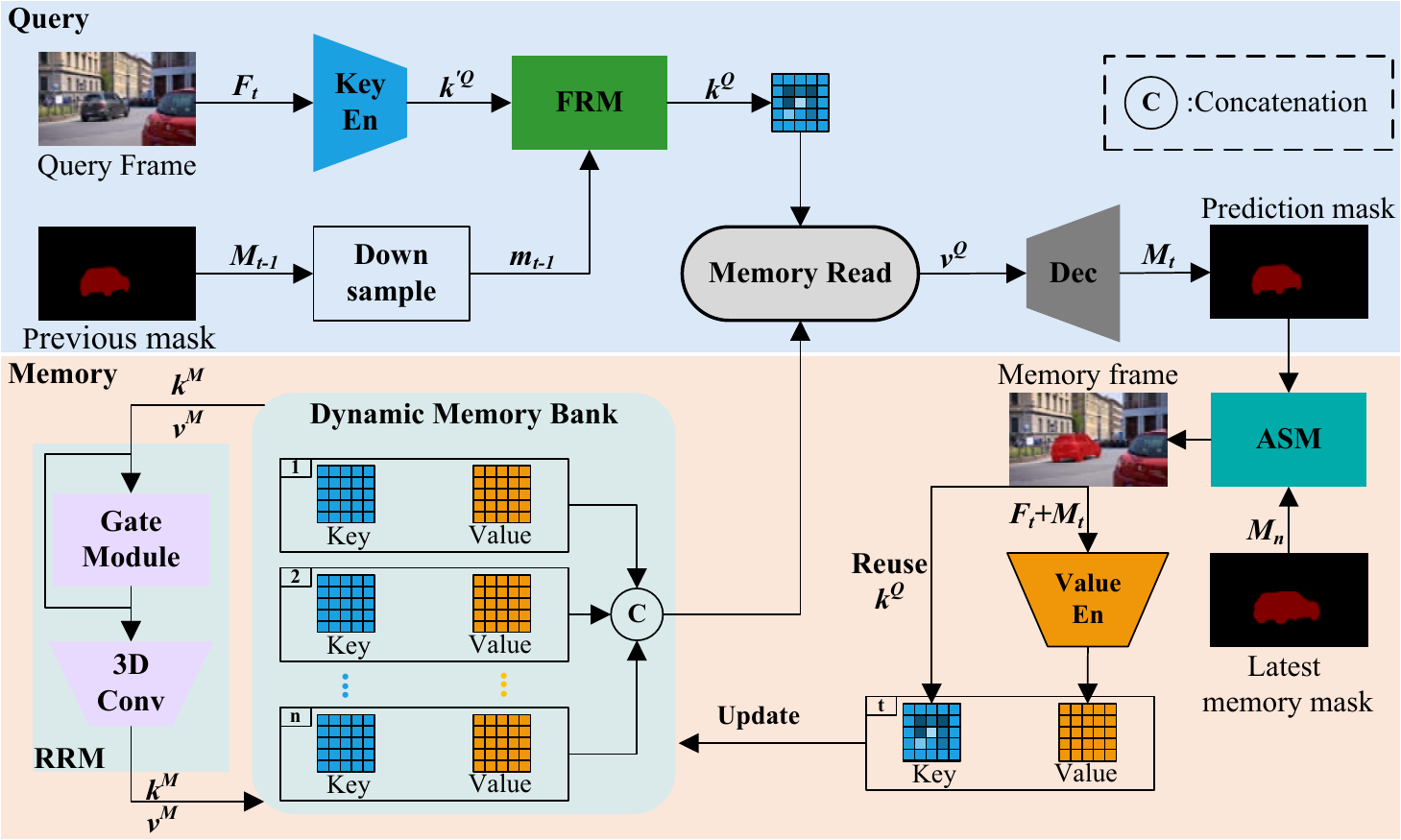}
	\caption{An overview of REMN. We follow the structure of STCN \cite{cheng2021rethinking} and propose a Foreground Reinforcing Module(FRM) for background distraction. Adaptive Sampling Module(ASM) is used to decide whether to save query frames as memory frames and Redundancy Reduction Module(RRM) to remove redundant features from the memory bank.}
	\label{Framework}
\end{figure*}
An overview of our proposed REMN appears in Fig. \ref{Framework}. Like STCN\cite{cheng2021rethinking}, we take the current frame as the query frame $F_{t}\in\mathbb{R}^{ H^{0} \times W^{0}\times 3}$ and the past frames with masks as memory frames, where $H^{0}$ and $W^{0}$ are the original size of the current frame. Key Encoder takes a frame as input to extract query key $k^{'Q}\in\mathbb{R}^{C^{k}\times H \times W}$ for query frame, where $H=\frac{H^{0}}{16}$ and $W=\frac{W^{0}}{16}$ are spatial dimensions, and ${C^{k}}$ is the dimension of the key space. For previous mask $M_{t-1}\in\mathbb{R}^{ H^{0} \times W^{0} \times 1}$, we downsample it to $m_{t-1}\in\mathbb{R}^{ H \times W \times 1}$. To make the query frame more robust against background distraction, the Foreground Reinforcing Module (FRM) provides an object location prior for $k^{'Q}$ using $M_{t-1}$ to obtain enhanced the query key $k^{Q}\in\mathbb{R}^{ H \times W \times C^{k}}$. 

The memory read component performs a non-local pixel-wise matching of $k^{Q}$ to memory key $k^{M}\in\mathbb{R}^{T \times H \times W \times C^{k}}$ of the memory frames. Then, it multiplies the matching result with memory value $v^{M}\in\mathbb{R}^{ T \times H \times W \times C^{v}}$  to aggregated readout feature $v^{Q}\in\mathbb{R}^{H \times W\times C^{v} }$ for query frame, where $T$ is the number of memory frames. The Decoder upsamples $v^{Q}$ to generate a prediction mask $M_{t}\in\mathbb{R}^{H^{0} \times W^{0}\times 1}$. 

The Adaptive Sampling Module (ASM) then computes the difference between $M_{t}$ and the latest memory frame mask $M_{n}\in\mathbb{R}^{H^{0} \times W^{0} \times 1}$ to decide whether to leave $F_{t}$ as a memory frame. Since all memory frames were once query frames, we reuse $k^{Q}$ for the memory key of the memory frame, followed by a Value Encoder which takes an image frame and a mask as input to extract the memory value. We also design a dynamic memory bank that can hold up to $N$ pairs of memory keys and values. Also, a Redundancy Reduction Module(RRM) adaptively removes redundant features from the memory bank when the number of memory frames exceeds $N$.
\subsection{Foreground Reinforcing Module}
\label{Section:TCM}
\begin{figure}[!t]
	\centering
	
	\includegraphics[width=0.45\textwidth]{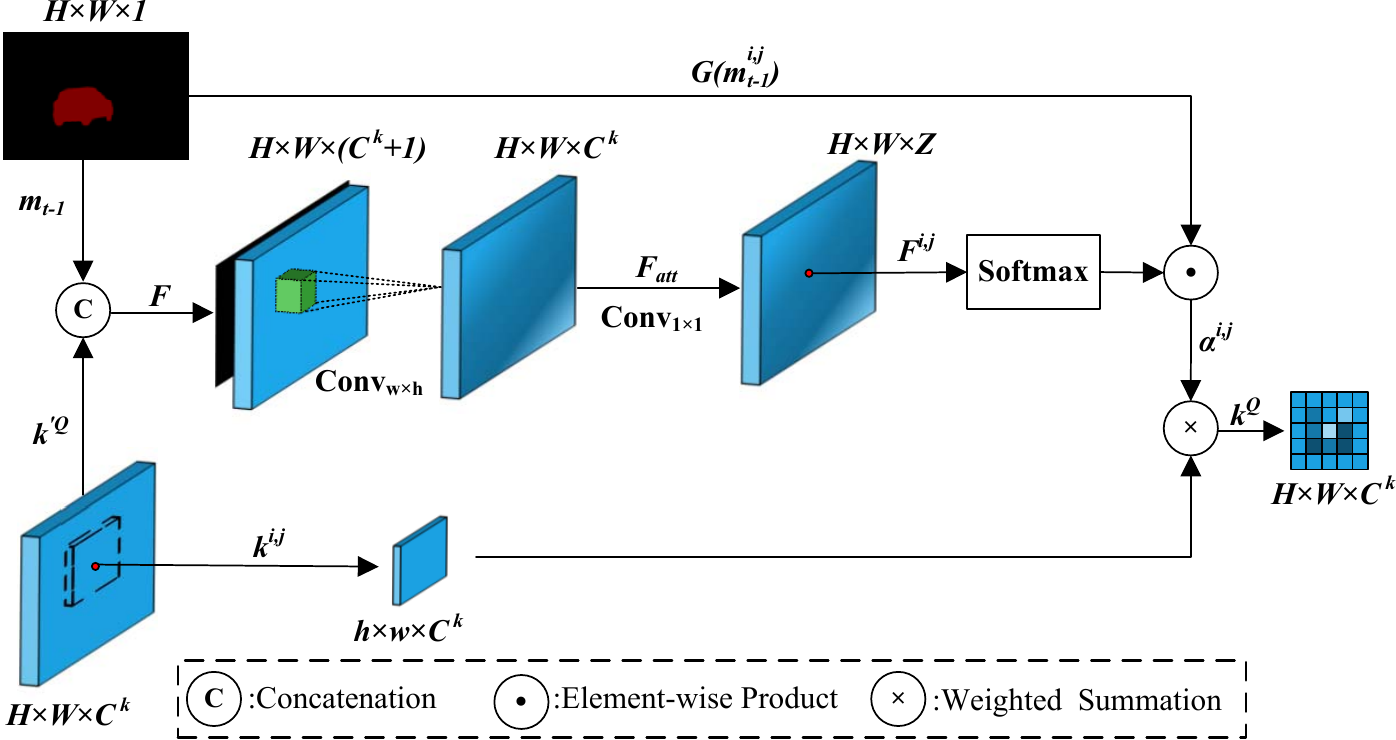}
	\caption{Implementation of Foreground Reinforcing Module.}
	\label{TCM}
\end{figure}

In this subsection, we introduce FRM as shown in Fig. \ref{TCM} that utilizes a local attention mechanism  to suppress background distraction due to the non-local pixel-wise matching. 

First, we concatenate $m_{t-1}$ with $k^{'Q}$ to obtain feature $F \in\mathbb{R}^{H\times W \times ( C^{k}+1)}$. Since the foreground object does not change much between adjacent frames, for each pixel in the previous frame, we only need to consider the spatial correlation between it and the neighborhoods of the query frame at the same location. Specifically, we restrict the receptive field of each pixel by a convolutional layer $Conv_{w \times h}$ with kernel size $w \times h$, thus generating a local attention feature $F_{att} \in\mathbb{R}^{H\times W \times  C^{k}}$. Next, the dimension of $F_{att}$ is transformed to $H \times W \times Z$ by a $1\times1$ convolutional layer $Conv_{1\times1}$, where $Z = w\times h$.

The vector of each pixel $(i,j)$ in $F_{att}$ is denoted as $F^{i,j}\in\mathbb{R}^{Z}$, which the softmax function normalized to generate local attention weights $\alpha^{i,j}\in\mathbb{R}^{Z}$. In addition, we reduce the weights of background pixels using the previous frame mask value $m_{t-1}^{i,j}$ at the corresponding position. The calculation is defined as :
\begin{equation}\label{sec 3.2-2}
	\alpha^{i,j}_{p} = \frac{\exp(F^{i,j}_{p})}{\sum_{m=1}^{Z}\exp(F^{i,j}_{m} )} \odot G(m_{t-1}^{i,j})
\end{equation}

\noindent where $p \in \left\{1,\cdots,Z\right\}$ and weight $\alpha^{i,j}_{p}$ is the spatial correlation at the $p^{th}$ pixel in the corresponding local region of query frame with respect to pixel $(i,j)$. $\odot$ represents the element-wise product. $G(\cdot)=\frac{\exp(\cdot)}{e}$ makes the weights of  background pixels tend to $0$ since the previous mask is not certainly correct. Finally, it obtains the enhanced query key $k^{Q}$ by weighted summation of each weight and its corresponding local query feature $k^{i,j}\in\mathbb{R}^{h \times w\times C}$. We depict the process as follows:
\begin{equation}\label{sec 3.2-3}
	k^{Q} = \sum\limits_{i=1}\limits^{H}\sum\limits_{j=1}\limits^{W}\alpha^{i,j} k^{i,j}
\end{equation}
\subsection{Adaptive Sampling Module}
\label{Section:DSM}
In this subsection, we introduce ASM, which can adaptively decide the sampling interval according to foreground object variation, which differs from previous works that used a fixed interval to save a query frame as memory frame. We decide whether to update the memory bank by computing the difference between the query frame and the latest memory frame for each foreground object. Specifically, for prediction mask $M_{t}^{i}$ of the $i^{th}$ object and its latest memory mask $M_{n}^{i}$, ASM directly utilize the intersection of union between them to measure the difference in position and shape of the foreground object. The process is depicted  as: 
\begin{equation}
	D^{i} = 1 - \frac{M_{t}^{i} \cap M_{n}^{i}}{M_{t}^{i} \cup M_{n}^{i}}
\end{equation}

\noindent where $D^{i}$ represents the rate of variation of the $i^{th}$ object. Once the rate of variation  of any object exceeds the threshold $\sigma$, ASM considers the query frame as a new memory frame and $M_{t}$ replaces $M_{n}$ as the latest memory mask for subsequent comparisons.
\subsection{Redundancy Reduction Module}
\label{Section:RRM}
In practice, more than controlling the sampling interval is necessary to reduce temporal redundancy in the memory bank. For this reason, we introduce a dynamic memory bank consisting of RRM that can dynamically remove redundant features and keep only the non-redundant parts. The complexity of memory reading in the previous memory-based methods increases exponentially with the number of memory frames. Unlike those methods, our proposed RRM not only removes redundant features in the memory bank but also keeps this complexity in a suitable range. 

Following STCN\cite{cheng2021rethinking}, for each memory frame, we store two items: the memory key and the memory value. When the number of memory frames reaches $N$ (i.e., $T=N$), the soft modulation gate takes $k^{M}$ as input and learns a probability vector $Prob\in\mathbb{R}^{S^{t}}$ for determining the number of features to be retained along the temporal dimension. $S^{t}$ is the search space size (i.e., the number of temporal policies). This process we summarize as follows:
\begin{equation}
	Prob = \omega(\mathcal{F}(\beta,\delta(\mathcal N(\mathcal{F}(\lambda,\mathcal A(k^{M})))))+\gamma)
\end{equation}
\noindent where $\mathcal A$ denotes Global Average Pooling(GAP) whose output size is $T \times 1 \times 1 \times C^{k}$, $\mathcal{F}(\cdot,\cdot)$ is a convolutional function, $\mathcal N$ indicates batch normalization, and $\lambda$, $\beta$ and $\theta$ are convolutional parameters. Both $\delta$ and $\omega$ represent  activation functions, in which $\delta$ is $Relu$ activation and $\omega$ is defined as:
\begin{equation}
	\omega(\cdot) =\max(0,Tanh(\cdot))
\end{equation}

\noindent With the proposed activation function $\omega$, $Prob$ range is $[0,1)$ which represents the probability of choosing each temporal policy. The index $S$ of the maximum value in $Prob$ is then found by the $Argmax$ function, which is the temporal policy output by the soft modulation gate. 

Further, for $k^{M}$ we use a 3D convolution with temporal stride of $2^{S+1}$, which can convert the size of $k^{M}$ from $T \times H \times W \times C^{k}$ to $\frac{T}{2^{S+1}}\times H \times W \times C^{k}$. Similarly, we perform this operation for $v^{M}$ and change its size to $\frac{T}{2^{S+1}}\times H \times W \times C^{v}$. 

To ensure that features output by RRM can be segmented with high quality, moreover, we introduce the RRM loss function as:
\begin{equation}
	\begin{split}
		\mathcal{L}_{RRM} = KL(k^{M} \Vert RRM(k^{M} )) + KL(v^{M} \Vert RRM(v^{M} ))
	\end{split}
\end{equation}

\noindent where $RRM(\cdot)$ denotes the operation of removing redundant features and $KL(\cdot \Vert \cdot)$ means Kullback-Leibler divergence. We make the distribution of RRM output similar to that of original features by $\mathcal{L}_{RRM}$ to avoid the loss of key features and poor segmentation.
\subsection{Memory Read and Decoder}
\label{Section:Read and Decoder}
As in STCN\cite{cheng2021rethinking}, we represent the non-local pixel-wise matching by an affinity matrix $W$, which is a softmax-normalized L2 similarity between $k^{M}$ and $k^{Q}$. The calculation is defined as:
\begin{equation}
	W(i,j)=\frac{\exp(- \Vert k^{M}_{i}-k^{Q}_{j}\Vert ^{2}_{2})}{\sum_{i}\exp(- \Vert k^{M}_{i}-k^{Q}_{j}\Vert ^{2}_{2})}
\end{equation}

\noindent where $i$ and $j$ are the positions of the memory key and query key. $- \Vert\cdot\Vert^{2}_{2}$ is the negative squared Euclidean distance, which we use to denote L2 similarity. $W$ can retrieve retrieving the most similar pixel feature location in the memory bank to the query frame. Finally, it is multiplied with $v^{M}$ in the memory bank to obtain the aggregated readout feature $v^{Q}$:
\begin{equation}
	v^{Q} = \sum\limits_{i} W(i,j)\odot v^{M}_{i}
\end{equation}

The structure of our Decoder is similar to the STM\cite{oh2019video} decoder, which consists of a residual block and two upsampling blocks. The Decoder takes $v^{Q}$ from the memory read as input, and the low-level features of the Key Encoder are skip-connected in the upsampling blocks to generate the predicted mask.

\section{Experiments}
In this section, we evaluate our proposed REMN on DAVIS 2017\cite{pont20172017}, YouTube-VOS 2018\cite{xu2018youtube}, and a long-time dataset proposed in \cite{cheng2021modular}. The common evaluation metrics are region similarity $\mathcal{J}$ and contour accuracy $\mathcal{F}$. To be more intuitive, we use their mean values $\mathcal{J\&F}$ for comparison. Furthermore, the inference speed also is a non-negligible metric, especially in the real world, and we use frames-per-second ($FPS$) to represent the inference speed of the network.
\subsection{Quantitative Comparison}\label{Quantitative Cpmparsion}
As shown in Table \ref{DVIAS2017}, REMN obtains competitive results on DAVIS 2017 with the fastest inference speed (i.e., $FPS$ of 27.5) compared to previous methods. On the validation set, REMN performs 1.0\% better on $\mathcal{J\&F}$ than our baseline STCN\cite{cheng2021rethinking}, and the inference speed is 36\% faster. Furthermore, GC\cite{li2020fast} which is most similar to our speed ($FPS$ 27.5 vs 25.0) has 14.9\% lower mean accuracy than ours. In particular, REMN consumes fewer computing resources while maintaining high accuracy, as shown in Fig. \ref{issues}(b). AFB-URR\cite{liang2020video} and GC\cite{li2020fast} consume constant cost during the inference as our network, but they become less effective when the foreground context is losing track. In addition, REMN also scores the highest $\mathcal{J}$ of 76.5\% on the test-dev set.
\begin{table}[htbp]
	\caption{Quantitative comparison on DAVIS 2017 validation and test-dev set. Bold and underline respectively means the best and second-best result in each column.Gen-error report the generalization gap between validation-set and test-dev set.
	 \label{DVIAS2017}}
	\centering
	\resizebox{0.48\textwidth}{!}{
		\begin{tabular}{cccccccccc}
			\hline
			\multicolumn{1}{c}{\multirow{2}*{Method}}&\multicolumn{3}{c}{Validation-set}&&\multicolumn{3}{c}{Test-dev set}& \multirow{2}*{$FPS$}& \multirow{2}*{Gen-error}\\
			\cline{2-4} 
			\cline{6-8}
			&$\mathcal{J\&F}$&$\mathcal{J}$&$\mathcal{F}$&&$\mathcal{J\&F}$&$\mathcal{J}$&$\mathcal{F}$&&\\
			\hline
			STM\cite{oh2019video}&81.8&79.2&84.3&&72.2&69.3&75.2&6.3&9.5\\
			AFB-URR\cite{liang2020video}&76.9&74.4&79.3&&-&-&-&6.8&-\\
			KMN\cite{seong2020kernelized}&82.8&80.0&85.6&&77.2&74.1&80.3&8.3&\underline{5.6}\\
			
			RMNet\cite{xie2021efficient}&83.5&81.0&86.0&&75.0&71.9&78.1&4.4&8.5\\
			CFBI\cite{yang2020collaborative}&81.9&79.1&84.6&&74.8&71.1&78.5&5.9&7.1\\
			GC\cite{li2020fast}&71.4&69.3&73.5&&-&-&-&\underline{25.0}&-\\
			MiVOS\cite{cheng2021modular}&84.5&81.7&87.4&&78.6&74.9&\underline{82.2}&11.2&6.0\\
			STCN\cite{cheng2021rethinking}&\underline{85.3}&\underline{82.0}&\underline{88.6}&&\textbf{79.9}&\underline{76.3}&\textbf{83.5}&20.2&\textbf{5.4}\\
			\hline
			
			RENM (Ours)&\textbf{86.3}&\textbf{83.4}&\textbf{89.2}&&\underline{79.2}&\textbf{76.5}&81.9&\textbf{27.5}&7.1\\
			\hline
	\end{tabular}}
\end{table}

We divide foreground objects in YouTube-VOS into two categories, including seen and unseen during the training stage, and measure the generalization ability of the network between categories. Table \ref{YouTube2018} shows the results of a quantitative comparison of the YouTube 2018 validation set between REMN and previous methods. In summary, REMN achieves a state-of-the-art performance of average accuracy and inference speed but also minimizes the generalization error.
\begin{table}[htbp]
	\caption{Quantitative comparison on YouTube 2018 validation set. $\mathcal{G}$ indicates overall mean. Gen-error report the generalization gap between seen and unseen categories.\label{YouTube2018}}
	\centering
	\resizebox{0.48\textwidth}{!}{
		\begin{tabular}{ccccccccc}
			\hline
			\multicolumn{1}{c}{\multirow{2}*{Method}}&\multicolumn{2}{c}{Seen}&&\multicolumn{2}{c}{Unseen}&\multicolumn{1}{c}{\multirow{2}*{$\mathcal{G}$}}& \multirow{2}*{$FPS$}&\multicolumn{1}{c}{\multirow{2}*{Gen-error}}\\
			\cline{2-3} 
			\cline{5-6}
			&$\mathcal{J}$&$\mathcal{F}$&&$\mathcal{J}$&$\mathcal{F}$&&&\\
			\hline
			STM\cite{oh2019video}&79.7&84.2&&72.8&80.9&79.4&-&5.1\\
			AFB-URR\cite{liang2020video}&78.8&83.1&&74.1&82.6&79.6&-&2.6\\
			KMN\cite{seong2020kernelized}&80.4&84.5&&73.8&81.4&80.0&-&4.9\\
			
			RMNet\cite{xie2021efficient}&82.1&85.7&&75.7&82.4&81.5&-&4.9\\
			CFBI\cite{yang2020collaborative}&80.6&85.1&&75.2&83.0&81.0&3.4&3.8\\
			GC\cite{li2020fast}&72.6&75.6&&68.9&75.7&73.2&-&\textbf{1.9}\\
			
			MiVOS\cite{cheng2021modular}&81.1&85.6&&77.7&86.2&82.7&-&\underline{2}\\
			
			STCN\cite{cheng2021rethinking}&\underline{83.2}&\underline{87.9}&&\underline{79.0}&\underline{87.3}&\underline{84.3}&\underline{13.2}&2.4\\
			\hline
			RENM (Ours)&\textbf{84.5}&\textbf{88.1}&&\textbf{81.0}&\textbf{88.3}&\textbf{85.5}&\textbf{25.6}&\textbf{1.9}\\
			\hline
		\end{tabular}
	}
\end{table}

To verify the performance of our network in long video scenarios, we use a long video dataset including 3 video sequences with an average length of over 2000 per video sequence. However, most of existing methods have unlimited memory banks, thus risking GPU memory overflow as the video length increases. We make them able to handle long-time videos by limiting their memory update frequency, but this also leads to decreased performance of these methods. As shown in Table \ref{long-time video dataset}, the closest match to our results is STCN, our network improves the accuracy by 0.6\% and is 3.5 times faster (FPS 25.3 vs. 7.2). We also extend the length of this long-time video by 3 times through replaying it, at which point the FPS of REMN is about 8 times (FPS 23.9 vs. 3.0) that of STCN. In summary, REMN performs competitive results on the long-time video dataset with an impressive inference speed.

\begin{table}[htbp]
	\caption{Quantitative comparison on the Long-time video dataset. $\times n$ means a video has n times the number of frames. $\times1\to\times3$ denotes the value differences.\label{long-time video dataset}}
	\centering
	\resizebox{0.48\textwidth}{!}{
		\begin{tabular}{ccccccccc}
			\hline
			\multicolumn{1}{c}{\multirow{2}*{Method}}&\multicolumn{2}{c}{Video Length($\times1$)}&&\multicolumn{2}{c}{Video Length($\times3$)}&&\multicolumn{2}{c}{$\times1\to\times3$}\\
			\cline{2-3} 
			\cline{5-6}
			\cline{8-9}
			&$\mathcal{J\&F}$&$FPS$&&$\mathcal{J\&F}$&$FPS$&&$\mathcal{J\&F}$&$FPS$\\
			\hline
			STM\cite{oh2019video}&80.6&-&&75.3&-&&-5.3&-\\
			AFB-URR\cite{liang2020video}&83.7&-&&83.8&-&&\underline{0.1}&-\\
			CFBI\cite{yang2020collaborative}&53.5&-&&58.9&-&&\textbf{5.4}&-\\
			STCN\cite{cheng2021rethinking}&\underline{87.3}&\underline{7.2}&&\underline{84.6}&\underline{3.0}&&-2.7&\underline{-4.2}\\
			\hline
			REMN (Ours)&\textbf{87.9}&\textbf{25.3}&&\textbf{87.4}&\textbf{23.9}&& -0.5&\textbf{-1.4}\\
			\hline
		\end{tabular}
	}
\end{table}
\subsection{Qualitative Comparison}
\begin{figure*}[htbp]
	\centering
	\subfloat{
		\rotatebox{90}{~~~STCN}
		\begin{minipage}[t]{0.15\textwidth}
			\centering
			\includegraphics[width=1\textwidth]{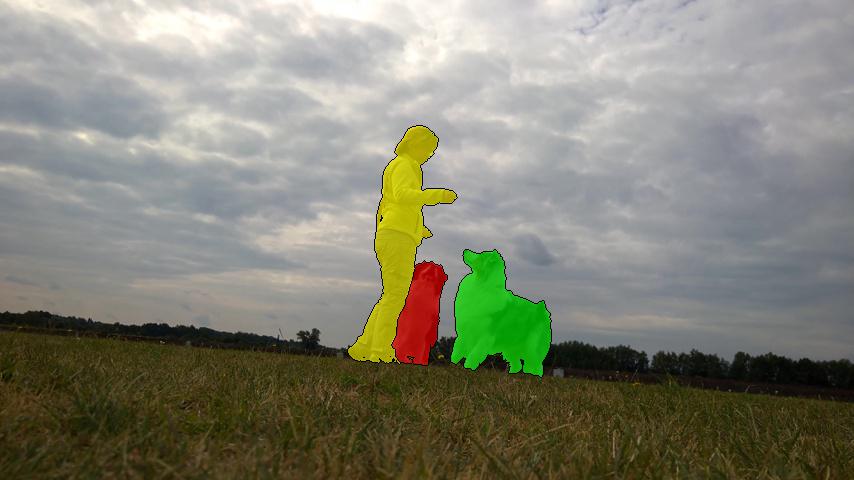}
		\end{minipage}
		\begin{minipage}[t]{0.15\textwidth}
			\centering
			\includegraphics[width=1\textwidth]{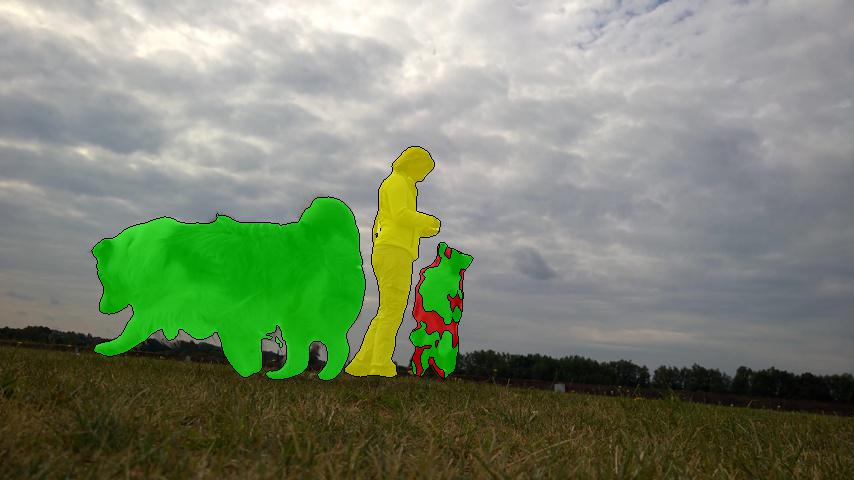}
		\end{minipage}
		\begin{minipage}[t]{0.15\textwidth}
			\centering
			\includegraphics[width=1\textwidth]{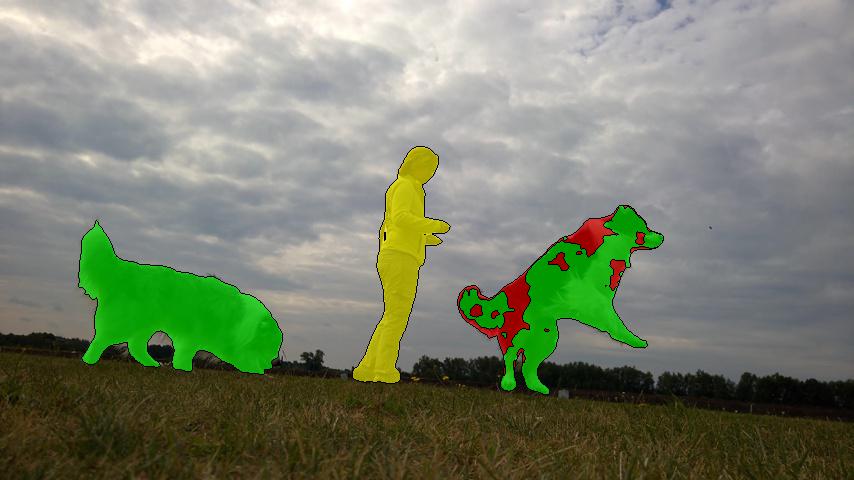}
		\end{minipage}
		\begin{minipage}[t]{0.15\textwidth}
			\centering
			\includegraphics[width=1\textwidth]{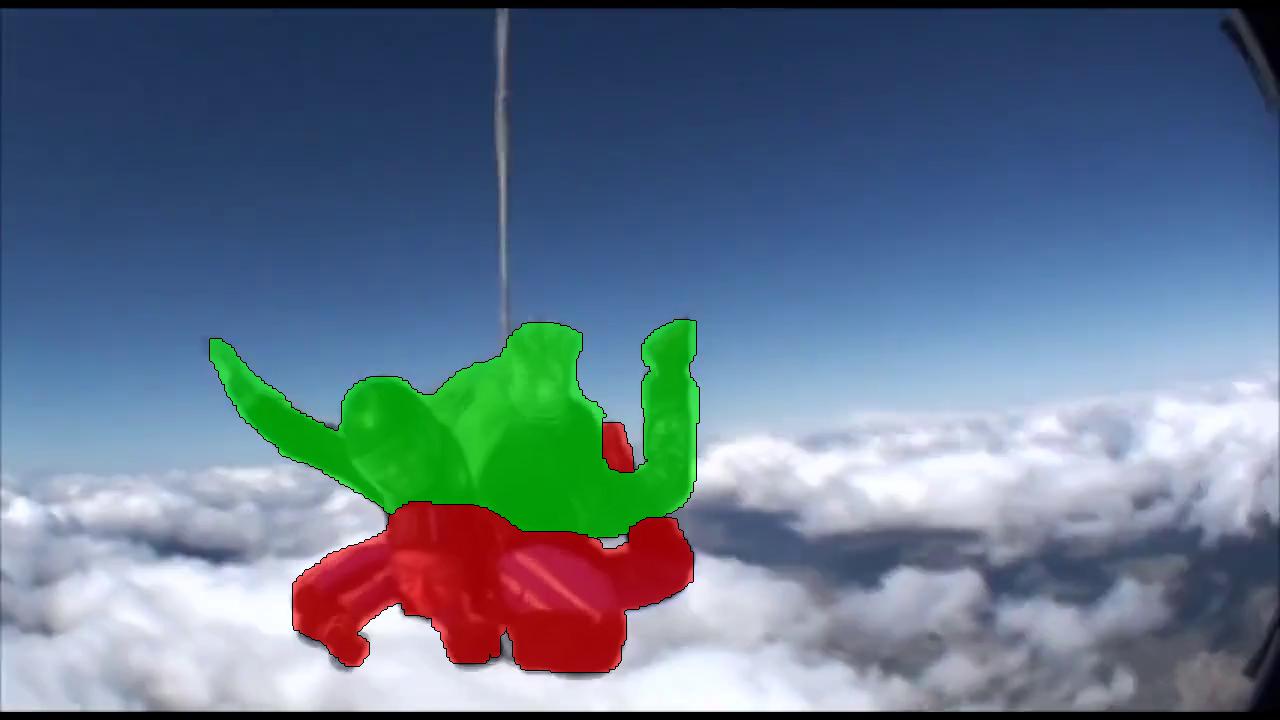}
		\end{minipage}
		\begin{minipage}[t]{0.15\textwidth}
			\centering
			\includegraphics[width=1\textwidth]{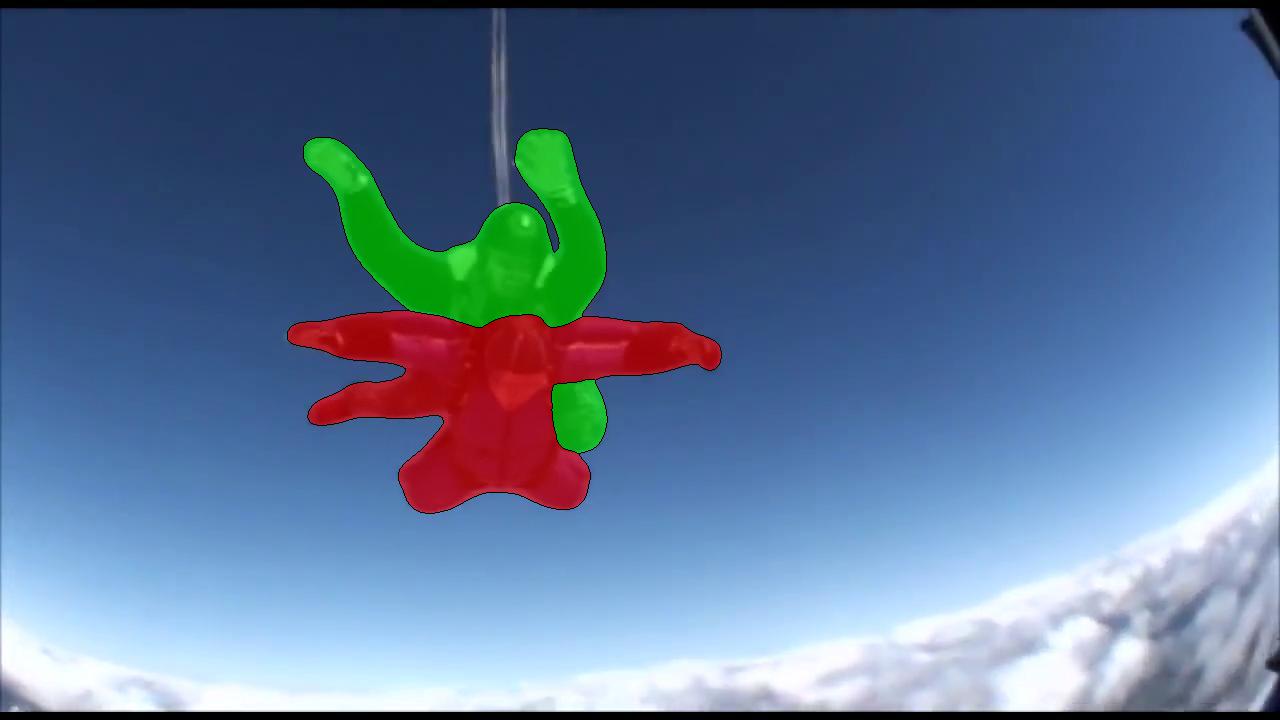}
		\end{minipage}
		\begin{minipage}[t]{0.15\textwidth}
			\centering
			\includegraphics[width=1\textwidth]{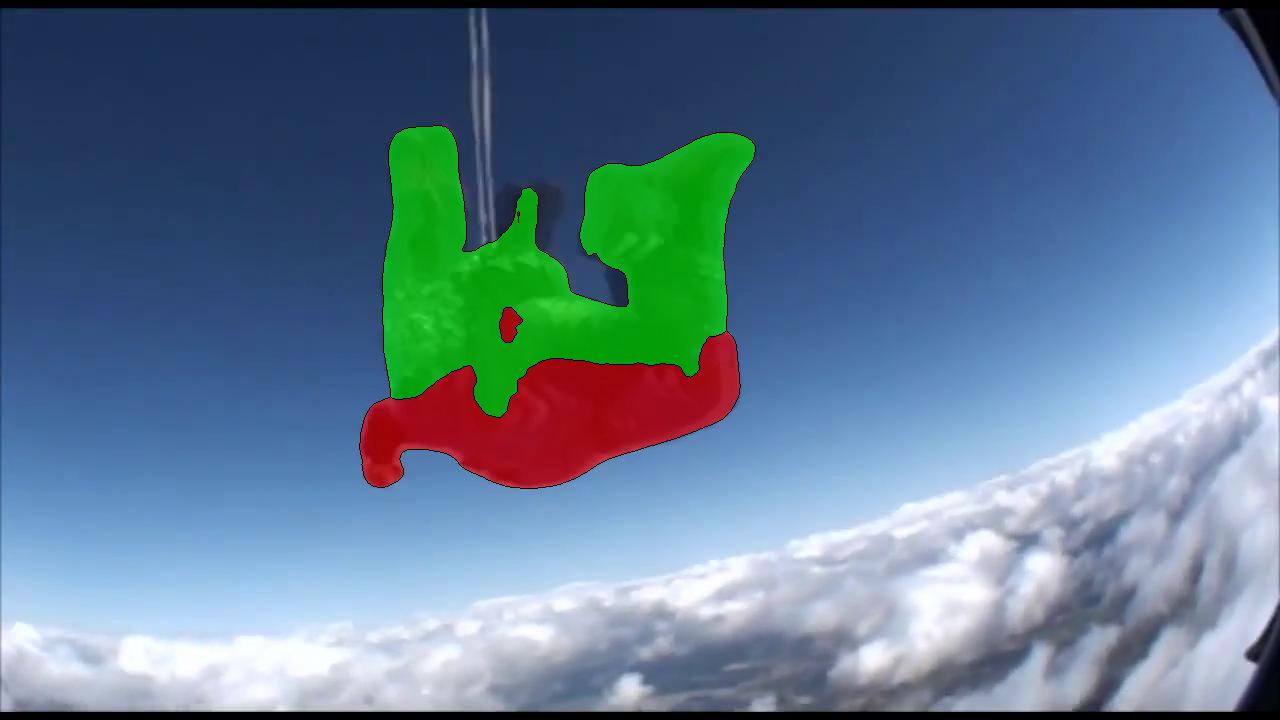}
		\end{minipage}
	}
	
	\subfloat{
		\rotatebox{90}{~~~REMN}
		\begin{minipage}[t]{0.15\textwidth}
			\centering
			\includegraphics[width=1\textwidth]{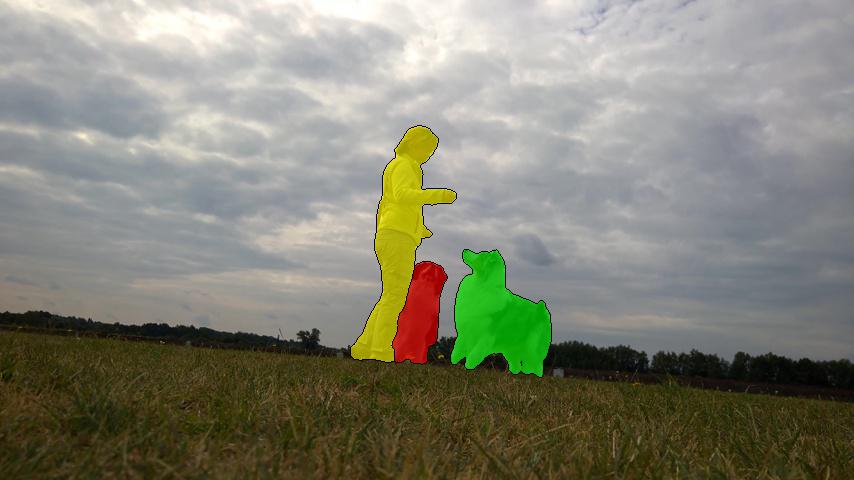}
		\end{minipage}
		\begin{minipage}[t]{0.15\textwidth}
			\centering
			\includegraphics[width=1\textwidth]{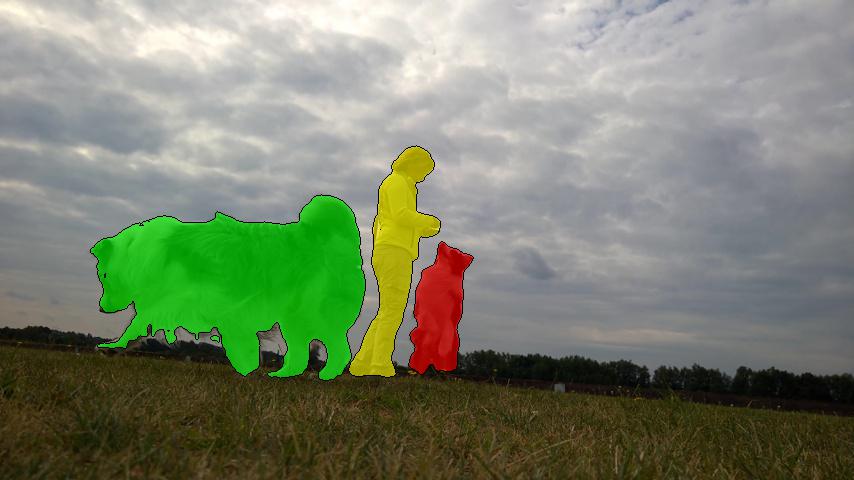}
		\end{minipage}
		\begin{minipage}[t]{0.15\textwidth}
			\centering
			\includegraphics[width=1\textwidth]{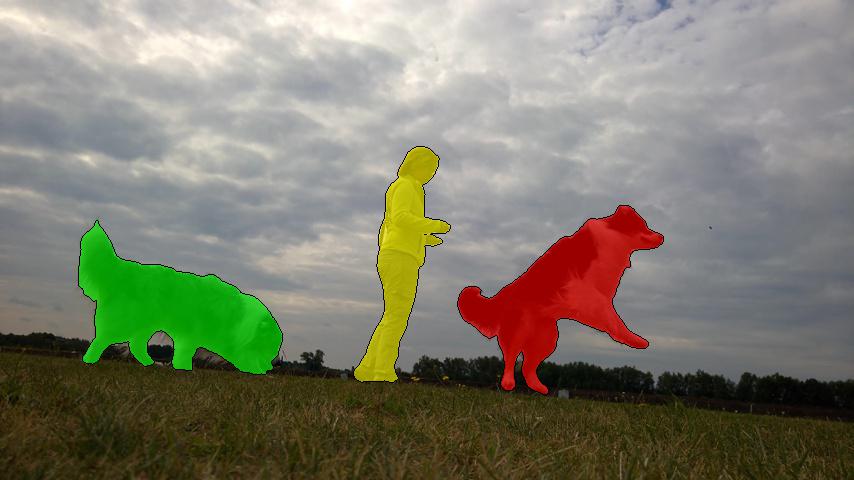}
		\end{minipage}
		\begin{minipage}[t]{0.15\textwidth}
			\centering
			\includegraphics[width=1\textwidth]{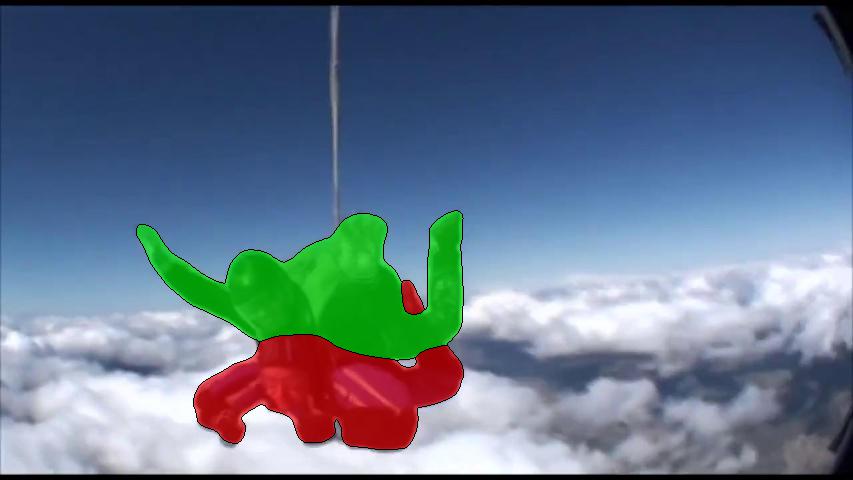}
		\end{minipage}
		\begin{minipage}[t]{0.15\textwidth}
			\centering
			\includegraphics[width=1\textwidth]{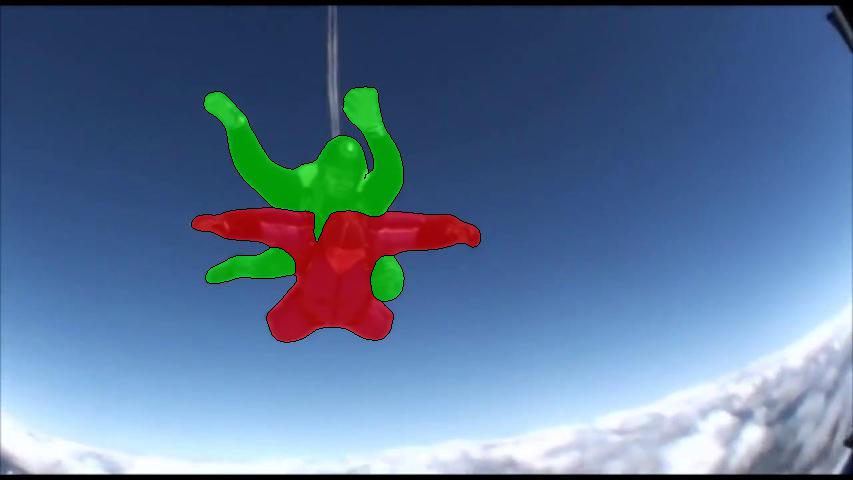}
		\end{minipage}
		\begin{minipage}[t]{0.15\textwidth}
			\centering
			\includegraphics[width=1\textwidth]{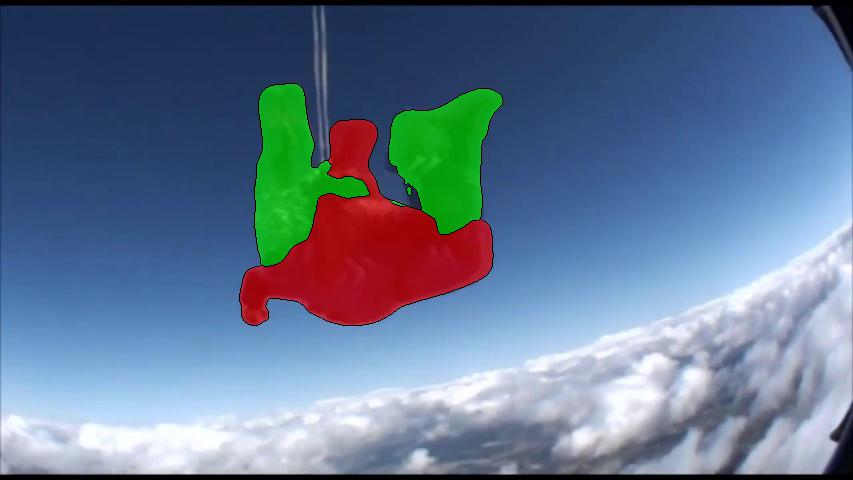}
		\end{minipage}
	}
	\caption{Qualitative results of REMN obtained by comparing it to baseline STCN\cite{cheng2021rethinking}.}
	\label{qc}
\end{figure*}

In Fig. \ref{qc}, we select two videos from DAVIS 2017 and YouTube 2018 as examples and visualize the results of qualitative comparison between REMN and baseline STCN\cite{cheng2021rethinking}. As can be seen, our proposed REMN has high robustness and achieves outstanding performance. 

\begin{figure}[htbp]
	\centering
	\subfloat[$F_{t}$]{
		\begin{minipage}{0.15\textwidth}
			\includegraphics[width=1\textwidth]{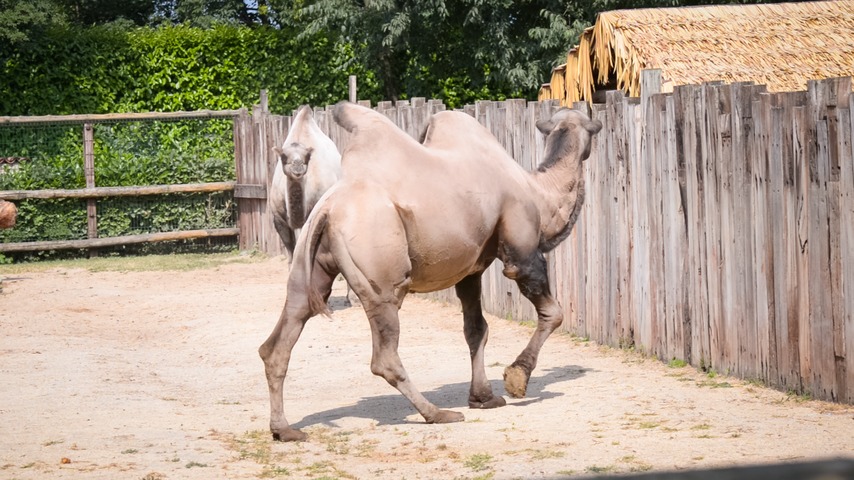}
	\end{minipage}}
	\subfloat[$M_{t-1}$]{
		\begin{minipage}{0.15\textwidth}
			\includegraphics[width=1\textwidth]{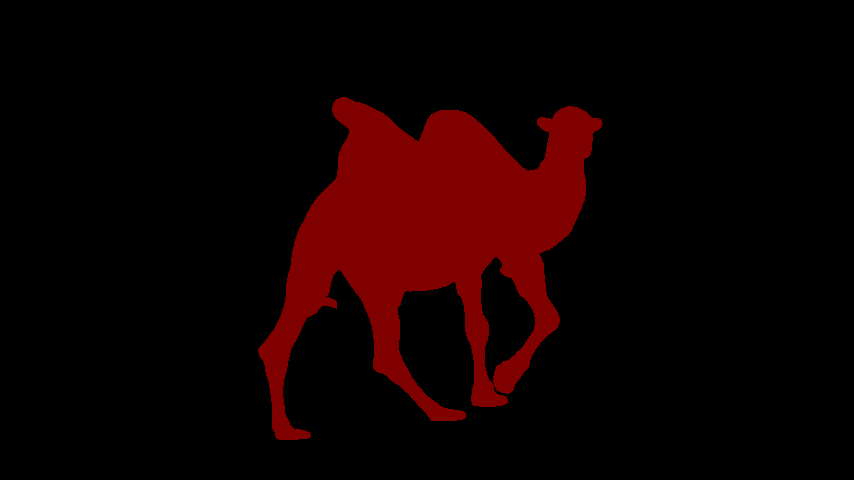}
	\end{minipage}}
	\subfloat[$k^{Q}$]{
		\begin{minipage}{0.15\textwidth}
			\includegraphics[width=1\textwidth,height=0.59in]{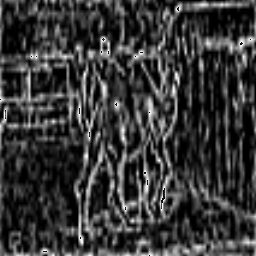}
	\end{minipage}}
	
	\subfloat[$F_{t}+M_{t}$]{
		\begin{minipage}{0.15\textwidth}
			\includegraphics[width=1\textwidth]{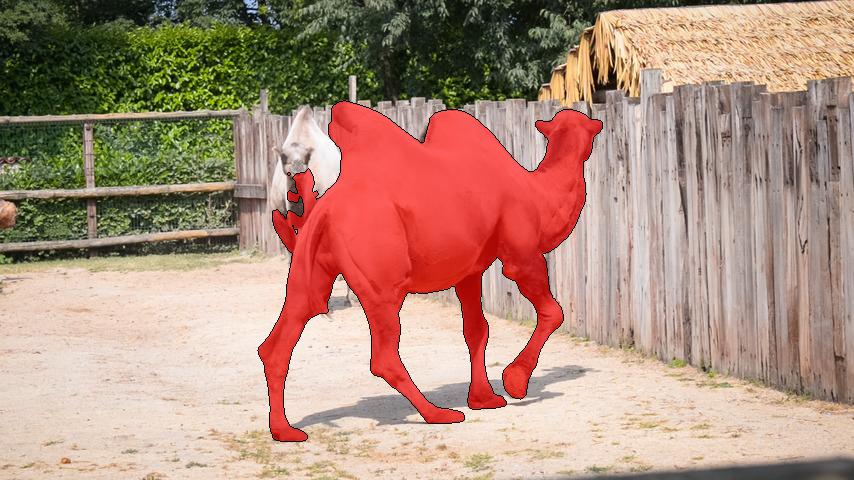}\vspace{2pt}
	\end{minipage}}
	\subfloat[$k^{'Q}$]{
		\begin{minipage}{0.15\textwidth}
			\includegraphics[width=1\textwidth,height=0.59in]{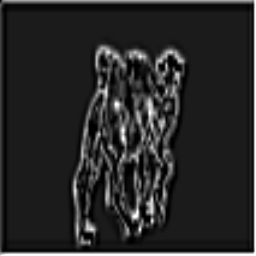}\vspace{2pt}
	\end{minipage}}
	\subfloat[$F_{t}+M^{'}_{t}$]{
		\begin{minipage}{0.15\textwidth}
			\includegraphics[width=1\textwidth]{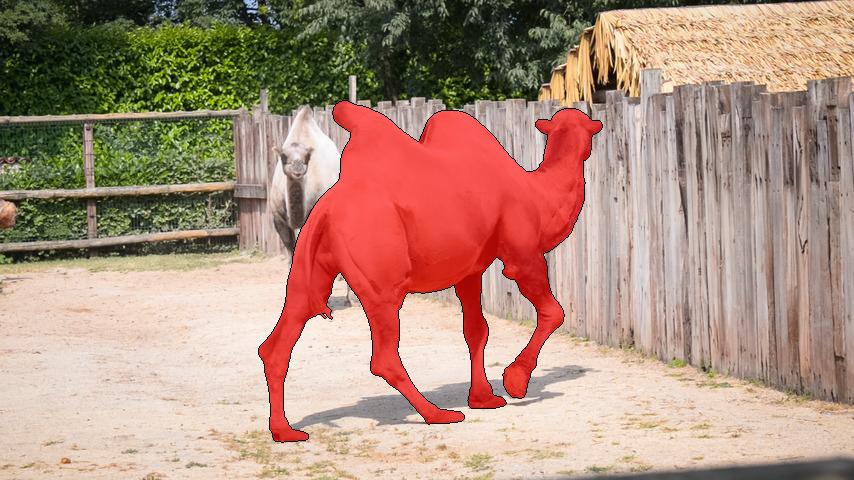}\vspace{2pt}
	\end{minipage}}
	\caption{The effectiveness of FRM. (a) The query frame. (b) Foreground object mask of the previous frame. (c) The original feature map produced by Key Encoder. (d) Result of using original feature. (e) The enhanced feature map produced by FRM. (f) Result of using enhanced feature.} 
	\label{The effectiveness of TCM}
\end{figure}
\begin{figure}[!h]
	\centering
	\subfloat[STCN\cite{cheng2021rethinking}]{
		\begin{minipage}{0.22\textwidth}
			\includegraphics[width=1\textwidth]{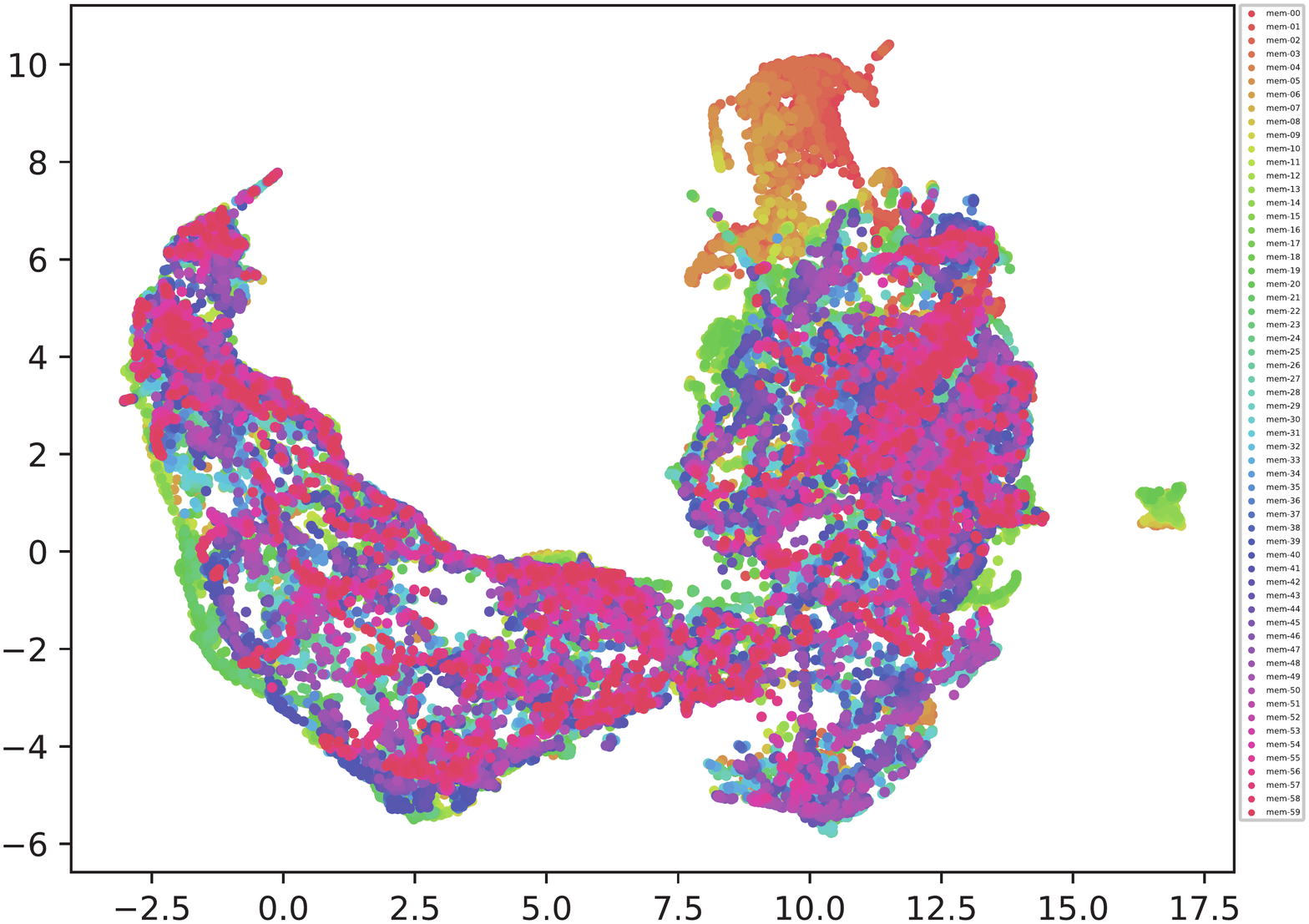}
		\end{minipage}
		\label{t-SNE-a}
	}
	\subfloat[Ours]{
		\begin{minipage}{0.22\textwidth}
			\includegraphics[width=1\textwidth]{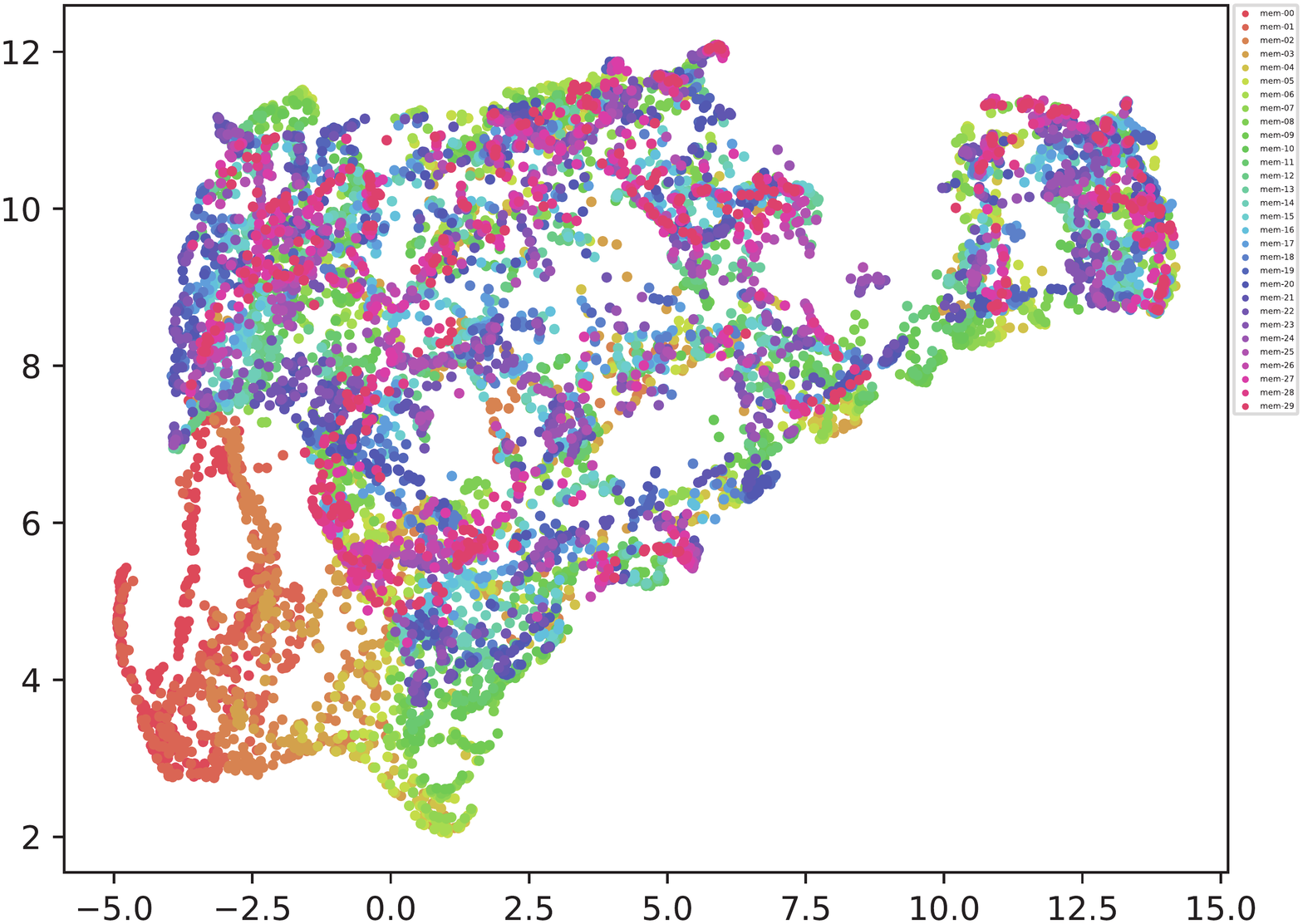}
		\end{minipage}
		\label{t-SNE-b}
	}
	\caption{The visualization results of our proposed dynamic memory bank compared to the baseline STCN\cite{cheng2021rethinking}. Different colors represent features of different memory frames.The more overlap between different color points indicates higher temporal redundancy.}
	\label{t-SNE}
\end{figure}

\subsection{Ablation Study}\label{Ablation Study}
We perform ablation experiments on the validation set of  DAVIS 2017 and YouTube 2018. Specifically, we report $\mathcal{J\&F}$ for DAVIS 2017 and $\mathcal{G}$ and $FPS$ for YouTube 2018, which denote the results of our network.

As shown in Table \ref{Module ablation study}, we perform ablation experiments on the three modules in REMN to analyze their effectiveness. When all modules are enabled (i.e., in the default configuration), they achieve $86.3\%$ and $85.5\%$ on DAVIS 2017 and YouTube 2018, and the inference speed reaches $25.6$ frames per second. When FRM is disabled, the performance drops $1.2\%$ and $1.4\%$, respectively. In Fig. \ref{The effectiveness of TCM}, we select a video sequence with background distraction to demonstrate the effectiveness of FRM. The visualization results show that the original features obtained by Key Encoder contain a background object, which can cause the background to be incorrectly segmented as foreground. In contrast, FRM avoids these erroneous segmentation cases by suppressing the background features. 

For memory management, we first use the strategy of sampling at 5 frame intervals instead of ASM, which reduces $FPS$ by $1.7$. We next deactivate RRM to maintain an unlimited memory bank. Although the results on YouTube 2018 improved by $0.2\%$ with this setting, $FPS$ is only $14.1$. Finally, we froze ASM and RRM (i.e., using the memory bank in STCN), lowering the inference speed to $12.9$ frames per second. This result is because STCN employs a memory bank that is unlimited and contains a large number of redundant features. To visualize the effectiveness of the dynamic memory bank consisting of ASM and RRM, we compared the distribution of features in the memory bank between STCN and ours. As shown in Fig. \ref{t-SNE}(a), many points overlap each other, so there is a high redundancy among the features in STCN's memory bank. On the contrary, the number of points in Fig. \ref{t-SNE}(b) is less than in \ref{t-SNE-a} and more evenly distributed, which indicates that our proposed ASM and RRM effectively remove redundant features.
\begin{table}[!t]
	\caption{Module ablation study.\label{Module ablation study}}
	\centering
	\begin{tabular}{cccccc}
		\hline
		FRM&ASM&RRM&$\mathcal{J\&F}$&$\mathcal{G}$&$FPS$\\
		\hline
		\checkmark&\checkmark&\checkmark&86.3&85.5&25.6\\
		\hline
		&\checkmark&\checkmark&85.1&84.1&25.8\\
		\checkmark&&\checkmark&86.3&85.4& 23.9\\
		\checkmark&\checkmark&&86.2&85.7&14.1\\
		\checkmark&&&85.6&84.7&12.9\\
		\hline 
	\end{tabular}
\end{table}

\subsection{Limitations}
We propose that RRM remove redundant features in the memory bank promptly, but some detailed features are inevitably lost. It will cause some foreground objects to not be segmented correctly, especially when the scale of foreground objects is small. We expect to obtain better performance in future work by refining the features of fine-scale foreground objects.

\bibliographystyle{IEEEtran.bst}
\bibliography{vos.bib}
%

\end{document}